\documentclass{article}

\usepackage{microtype}
\usepackage{lipsum}
\usepackage{graphicx}
\usepackage{rotating}
\usepackage{tabularx}
\usepackage{subcaption}
\usepackage{enumitem}
\usepackage{booktabs} 
\usepackage{hyperref}
\usepackage{math_macro}

\usepackage[accepted]{icml2018}

\icmltitlerunning{Junction Tree Variational Autoencoder for Molecular Graph Generation}

\begin{document}

\twocolumn[
\icmltitle{Junction Tree Variational Autoencoder for Molecular Graph Generation}



\icmlsetsymbol{equal}{*}

\begin{icmlauthorlist}
\icmlauthor{Wengong Jin}{mit}
\icmlauthor{Regina Barzilay}{mit}
\icmlauthor{Tommi Jaakkola}{mit}
\end{icmlauthorlist}

\icmlaffiliation{mit}{MIT Computer Science \& Artificial Intelligence Lab}

\icmlcorrespondingauthor{Wengong Jin}{wengong@csail.mit.edu}

\icmlkeywords{Deep Learning, Generative Model, Graph Generation}

\vskip 0.3in
]



\printAffiliationsAndNotice{}  

\begin{abstract}

We seek to automate the design of molecules based on specific chemical properties. In computational terms, this task involves continuous embedding and generation of molecular graphs. Our primary contribution is the direct realization of molecular graphs, a task previously approached by generating linear SMILES strings instead of graphs. Our \emph{junction tree variational autoencoder} generates molecular graphs in two phases, by first generating a tree-structured scaffold over chemical substructures, and then combining them into a molecule with a graph message passing network. This approach allows us to incrementally expand molecules while maintaining chemical validity at every step. We evaluate our model on multiple tasks ranging from molecular generation to optimization. Across these tasks, our model outperforms previous state-of-the-art baselines by a significant margin.

\end{abstract}
\section{Introduction}

The key challenge of drug discovery is to find target molecules with desired chemical properties. Currently, this task takes years of development and exploration by expert chemists and pharmacologists. Our ultimate goal is to automate this process. From a computational perspective, we decompose the challenge into two complementary subtasks: learning to represent molecules in a continuous manner that facilitates the prediction and optimization of their properties (encoding); and  learning to map an optimized continuous representation back into a molecular graph with improved properties (decoding). While deep learning has been extensively investigated for molecular graph encoding~\cite{duvenaud2015convolutional,kearnes2016molecular,gilmer2017neural}, the harder combinatorial task of molecular graph generation from latent representation remains under-explored. 

\begin{figure}[t]
\centering
\includegraphics[width=0.45\textwidth]{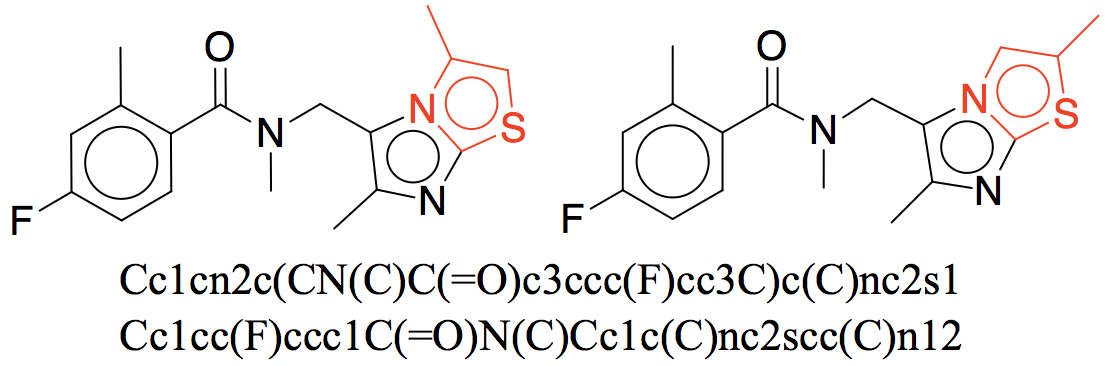}
\vspace{-8pt}
\caption{Two almost identical molecules with markedly different canonical SMILES in RDKit. The edit distance between two strings is 22 (50.5\% of the whole sequence).}
\label{fig:smiles}
\vspace{-12pt}
\end{figure}

Prior work on drug design formulated the graph generation task as a string generation problem~\citep{gomez2016automatic,kusner2017grammar} in an attempt to side-step direct generation of graphs. Specifically, these models start by generating SMILES~\citep{weininger1988smiles}, a linear string notation used in chemistry to describe molecular structures. SMILES strings can be translated into graphs via deterministic mappings (e.g., using RDKit~\citep{landrum2006rdkit}). 
However, this design has two critical limitations. First, the SMILES representation is not designed to capture molecular similarity. For instance, two molecules with similar chemical structures may be encoded into markedly different SMILES strings (e.g., Figure~\ref{fig:smiles}).
This prevents generative models like variational autoencoders from learning smooth molecular embeddings. 
Second, essential chemical properties such as molecule validity are easier to express on graphs rather than linear SMILES representations. We hypothesize that operating directly on graphs improves generative modeling of valid chemical structures.

\begin{figure}[t]
\centering
\includegraphics[width=0.45\textwidth]{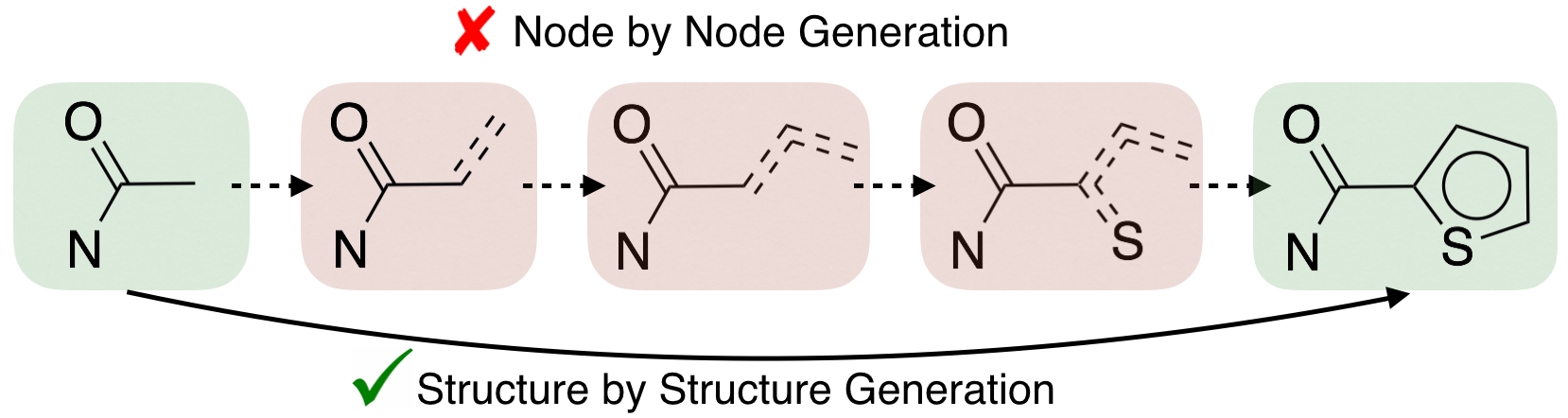}
\vspace{-10pt}
\caption{Comparison of two graph generation schemes: Structure by structure approach is preferred as it avoids invalid intermediate states (marked in red) encountered in node by node approach.}
\label{fig:structgen}
\vspace{-14pt}
\end{figure}

Our primary contribution is a new generative model of molecular graphs. While one could imagine solving the problem in a standard manner -- generating graphs node by node~\citep{li2018learning} -- the approach is not ideal for molecules. This is because creating molecules atom by atom would force the model to generate chemically invalid intermediaries (see, e.g., Figure~\ref{fig:structgen}), delaying validation until a complete graph is generated. Instead, we propose to generate molecular graphs in two phases by exploiting valid subgraphs as components. The overall generative approach, cast as a \emph{junction tree variational autoencoder}\footnote{https://github.com/wengong-jin/icml18-jtnn}, first generates a tree structured object (a junction tree) whose role is to represent the scaffold of subgraph components and their coarse relative arrangements. The components are valid chemical substructures automatically extracted from the training set using tree decomposition and are used as building blocks. In the second phase, the subgraphs (nodes in the tree) are assembled together into a molecular graph.

We evaluate our model on multiple tasks ranging from molecular generation to optimization of a given molecule according to desired properties. As baselines, we utilize state-of-the-art SMILES-based generation approaches~\cite{kusner2017grammar,dai2018syntax-directed}. We demonstrate that our model produces 100\% valid molecules when sampled from a prior distribution, outperforming the top performing baseline by a significant margin. In addition, we show that our model excels in discovering molecules with desired properties, yielding a 30\% relative gain over the baselines.

\section{Junction Tree Variational Autoencoder}
\label{sec:jtnn}

Our approach extends the variational autoencoder~\citep{kingma2013auto} to molecular graphs by introducing a suitable encoder and a matching decoder. 
Deviating from previous work~\citep{gomez2016automatic,kusner2017grammar}, we interpret each molecule as having been built from subgraphs chosen out of a vocabulary of valid components. 
These components are used as building blocks both when encoding a molecule into a vector representation as well as when decoding latent vectors back into valid molecular graphs. 
The key advantage of this view is that the decoder can realize a valid molecule piece by piece by utilizing the collection of valid components and how they interact, rather than trying to build the molecule atom by atom through chemically invalid intermediaries (Figure~\ref{fig:structgen}). An aromatic bond, for example, is chemically invalid on its own unless the entire aromatic ring is present. It would be therefore challenging to learn to build rings atom by atom rather than by introducing rings as part of the basic vocabulary.

Our vocabulary of components, such as rings, bonds and individual atoms, is chosen to be large enough so that a given molecule can be covered by overlapping components or \emph{clusters} of atoms. 
The clusters serve the role analogous to cliques in graphical models, as they are expressive enough that a molecule can be covered by overlapping clusters without forming cluster cycles. 
In this sense, the clusters serve as cliques in a (non-optimal) triangulation of the molecular graph. We form a junction tree of such clusters and use it as the tree representation of the molecule. 
Since our choice of cliques is constrained a priori, we cannot guarantee that a junction tree exists with such clusters for an arbitrary molecule. However, our clusters are built on the basis of the molecules in the training set to ensure that a corresponding junction tree can be found. Empirically, our clusters cover most of the molecules in the test set.

\begin{figure}[t]
\centering
\includegraphics[width=0.47\textwidth]{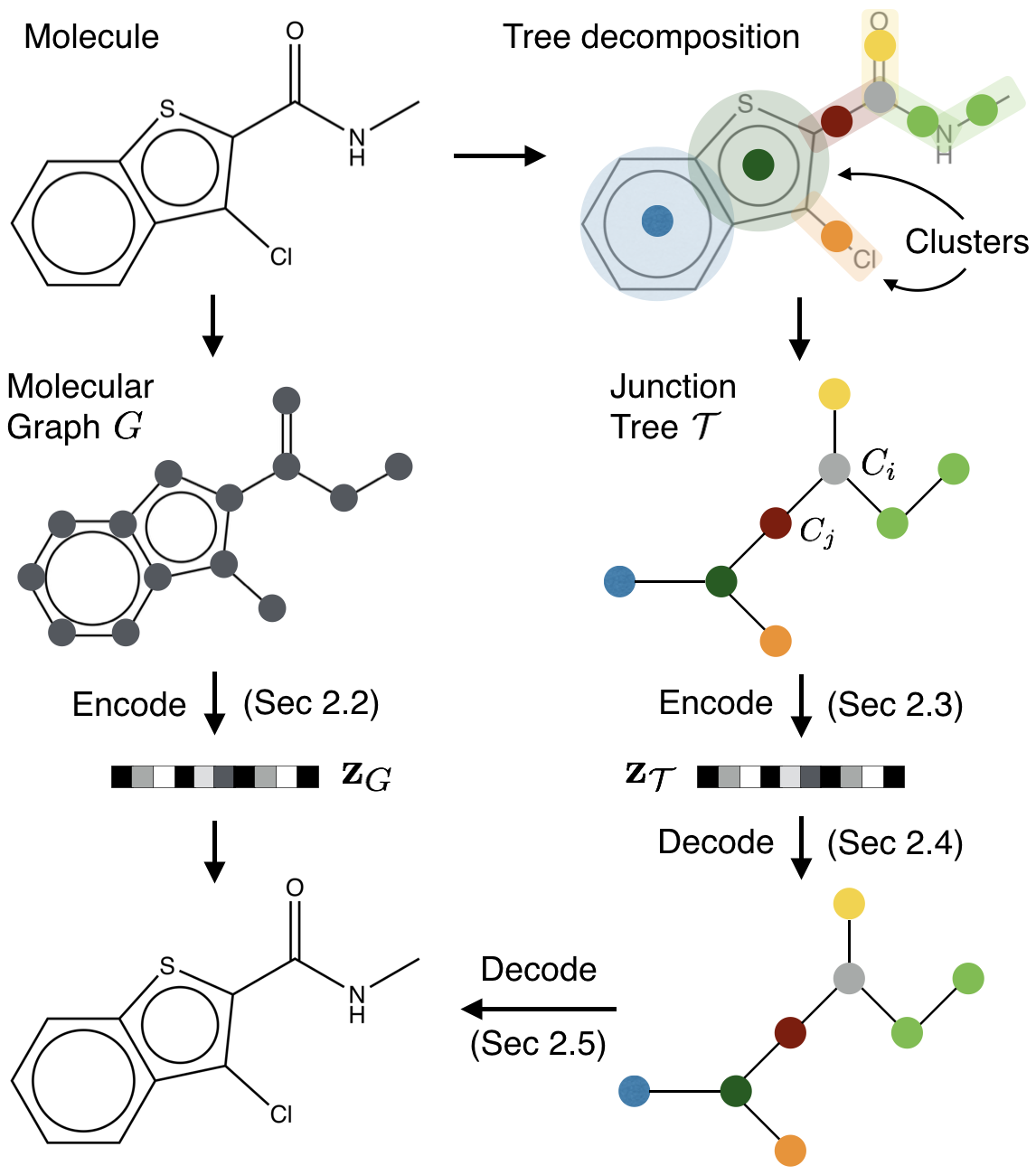}
\vspace{-8pt}
\caption{Overview of our method: A molecular graph $G$ is first decomposed into its junction tree $\tree_G$, where each colored node in the tree represents a substructure in the molecule. We then encode both the tree and graph into their latent embeddings $\z_\tree$ and $\z_G$. To decode the molecule, we first reconstruct junction tree from $\z_\tree$, and then assemble nodes in the tree back to the original molecule.}
\label{fig:paradigm}
\vspace{-13pt}
\end{figure}

The original molecular graph and its associated junction tree offer two complementary representations of a molecule. We therefore encode the molecule into a two-part latent representation $\z=[\z_\tree,\z_G]$ where $\z_\tree$ encodes the tree structure and what the clusters are in the tree without fully capturing how exactly the clusters are mutually connected. $\z_G$ encodes the graph to capture the fine-grained connectivity. Both parts are created by tree and graph encoders $q(\z_\tree | \tree)$ and $q(\z_G | G)$. The latent representation is then decoded back into a molecular graph in two stages. 
As illustrated in Figure~\ref{fig:paradigm}, we first reproduce the junction tree using a tree decoder $p(\tree | \z_\tree)$ based on the information in $\z_\tree$. Second, we predict the fine grain connectivity between the clusters in the junction tree using a graph decoder $p(G |\tree,\z_G)$ to realize the full molecular graph. The junction tree approach allows us to maintain chemical feasibility during generation. 

\textbf{Notation } A molecular graph is defined as $G=(V,E)$ where $V$ is the set of atoms (vertices) and $E$ the set of bonds (edges). Let $N(x)$ be the neighbor of $x$. We denote sigmoid function as $\sigma(\cdot)$ and ReLU function as $\tau(\cdot)$. We use $i,j,k$ for nodes in the tree and $u,v,w$ for nodes in the graph.

\subsection{Junction Tree}
A tree decomposition maps a graph $G$ into a \textit{junction tree} by contracting certain vertices into a single node so that $G$ becomes cycle-free. Formally, given a graph $G$, a junction tree $\tree_G=(\calV, \calE,\mathcal{X})$ is a connected labeled tree whose node set is $\calV = \{C_1,\cdots,C_n\}$ and edge set is $\calE$. Each node or \textit{cluster} $C_i=(V_i,E_i)$ is an induced subgraph of $G$, satisfying the following constraints:
\vspace{-3pt}
\begin{enumerate} \itemsep=1pt
\item The union of all clusters equals $G$. That is, $\bigcup_i V_i = V$ and $\bigcup_i E_i = E$.
\item Running intersection: For all clusters $C_i, C_j$ and $C_k$, $V_i \cap V_j \subseteq V_k$ if $C_k$ is on the path from $C_i$ to $C_j$. 
\end{enumerate}
\vspace{-3pt}
Viewing induced subgraphs as cluster labels, junction trees are labeled trees with label vocabulary $\mathcal{X}$. By our molecule tree decomposition, $\mathcal{X}$ contains only cycles (rings) and single edges. Thus the vocabulary size is limited ($\len{\mathcal{X}}=780$ for a standard dataset with 250K molecules).

\textbf{Tree Decomposition of Molecules } Here we present our tree decomposition algorithm tailored for molecules, which finds its root in chemistry~\citep{rarey1998feature}. Our cluster vocabulary $\mathcal{X}$ includes chemical structures such as bonds and rings (Figure~\ref{fig:paradigm}). Given a graph $G$, we first find all its simple cycles, and its edges not belonging to any cycles. Two simple rings are merged together if they have more than two overlapping atoms, as they constitute a specific structure called bridged compounds~\cite{claydenorganic}. Each of those cycles or edges is considered as a cluster. Next, a cluster graph is constructed by adding edges between all intersecting clusters. Finally, we select one of its spanning trees as the junction tree of $G$ (Figure~\ref{fig:paradigm}). 
As a result of ring merging, any two clusters in the junction tree have at most two atoms in common, facilitating efficient inference in the graph decoding phase. The detailed procedure is described in the supplementary.
\subsection{Graph Encoder}
We first encode the latent representation of $G$ by a graph message passing network~\cite{dai2016discriminative,gilmer2017neural}. Each vertex $v$ has a feature vector $\x_v$ indicating the atom type, valence, and other properties. Similarly, each edge $(u,v)\in E$ has a feature vector $\x_{uv}$ indicating its bond type, and two hidden vectors $\bnu_{uv}$ and $\bnu_{vu}$ denoting the message from $u$ to $v$ and vice versa. Due to the loopy structure of the graph, messages are exchanged in a loopy belief propagation fashion:
\begin{equation}
\bnu_{uv}^{(t)} = \tau(\W_1^g \x_u + \W_2^g \x_{uv} + \W_3^g \sum_{w \in N(u) \backslash v} \bnu_{wu}^{(t-1)})
\label{eq:mpn}
\end{equation}
where $\bnu_{uv}^{(t)}$ is the message computed in $t$-th iteration, initialized with $\bnu_{uv}^{(0)}=\mbf{0}$. After $T$ steps of iteration, we aggregate those messages as the latent vector of each vertex, which captures its local graphical structure:
\begin{equation}
\h_u = \tau(\U_1^g \x_u + \sum\nolimits_{v \in N(u)} \U_2^g \bnu_{vu}^{(T)})
\end{equation}
The final graph representation is $\h_G = \sum_i \h_i / \len{V}$. The mean $\bmu_G$ and log variance $\log \bsigma_G$ of the variational posterior approximation are computed from $\h_G$ with two separate affine layers. $\z_G$ is sampled from a Gaussian $\mathcal{N}(\bmu_G,\bsigma_G)$.

\subsection{Tree Encoder}
We similarly encode $\tree_G$ with a tree message passing network. 
Each cluster $C_i$ is represented by a one-hot encoding $\x_i$ representing its label type. Each edge $(C_i,C_j)$ is associated with two message vectors $\m_{ij}$ and $\m_{ji}$. We pick an arbitrary leaf node as the root and propagate messages in two phases. In the first bottom-up phase, messages are initiated from the leaf nodes and propagated iteratively towards root. In the top-down phase, messages are propagated from the root to all the leaf nodes. Message $\m_{ij}$ is updated as:
\begin{equation}
\m_{ij} = \GRU(\x_i, \{ \m_{ki}\}_{k \in N(i) \backslash j}) 
\end{equation}
where $\GRU$ is a Gated Recurrent Unit~\cite{chung2014empirical,li2015gated} adapted for tree message passing:
\begin{eqnarray}
\mbf{s}_{ij} &=& \sum\nolimits_{k \in N(i) \backslash j} \m_{ki} \\
\z_{ij} &=& \sigma(\W^z \x_i + \U^z \mbf{s}_{ij} + \mbf{b}^z) \\
\mbf{r}_{ki} &=& \sigma(\W^r \x_i + \U^r \m_{ki} + \mbf{b}^r) \\
\widetilde{\m}_{ij} &=& \tanh(\W \x_i + \U \sum_{k \in N(i) \backslash j} \mbf{r}_{ki} \odot \m_{ki}) \\
\m_{ij} &=& (1 - \z_{ij}) \odot \mbf{s}_{ij} + \z_{ij} \odot \widetilde{\m}_{ij}
\end{eqnarray}
The message passing follows the schedule where $\m_{ij}$ is computed only when all its precursors $\{ \m_{ki}\ \; | \; k \in N(i) \backslash j\}$ have been computed. This architectural design is motivated by the belief propagation algorithm over trees and is thus different from the graph encoder.

After the message passing, we obtain the latent representation of each node $\h_i$ by aggregating its inward messages:
\begin{equation}
\h_i = \tau(\W^o \x_i + \sum\nolimits_{k \in N(i)} \U^o \m_{ki})
\end{equation}
The final tree representation is $\h_{\tree_G} = \h_{root}$, which encodes a rooted tree $(\tree,root)$. Unlike the graph encoder, we do not apply node average pooling because it confuses the tree decoder which node to generate first. 
$\z_{\tree_G}$ is sampled in a similar way as in the graph encoder. For simplicity, we abbreviate $\z_{\tree_G}$ as $\z_\tree$ from now on.

This tree encoder plays \emph{two} roles in our framework. First, it is used to compute $\z_\tree$, which only requires the bottom-up phase of the network. Second, after a tree $\widehat{\tree}$ is decoded from $\z_\tree$, it is used to compute messages $\widehat{\m}_{ij}$ over the entire $\widehat{\tree}$, to provide essential contexts of every node during graph decoding. This requires both top-down and bottom-up phases. We will elaborate this in section~\ref{sec:gdecode}.
\subsection{Tree Decoder}
\begin{figure}[t]
\centering
\includegraphics[width=0.48\textwidth]{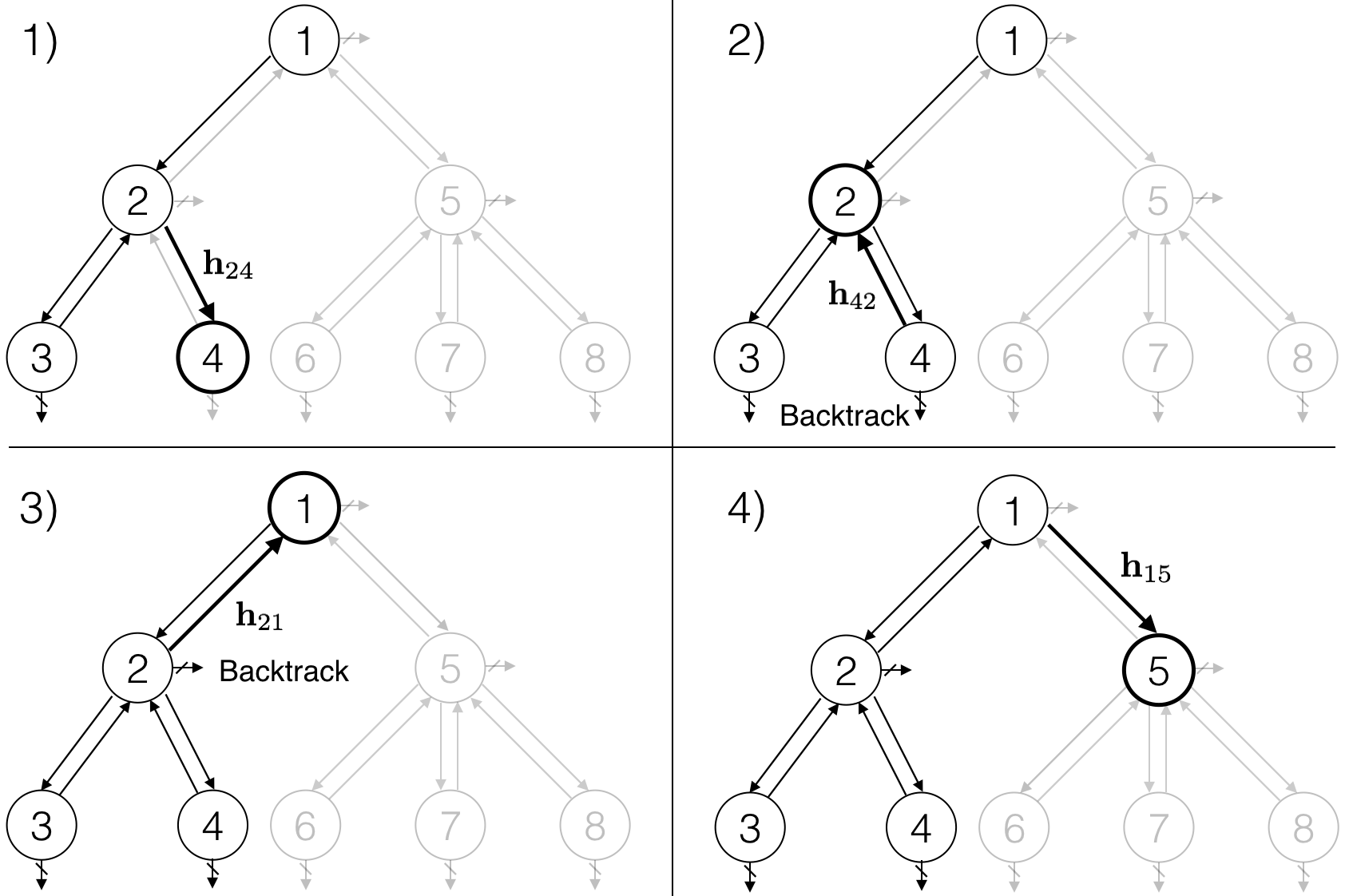}
\vspace{-15pt}
\caption{Illustration of the tree decoding process. Nodes are labeled in the order in which they are generated. 1) Node 2 expands child node 4 and predicts its label with message $\h_{24}$. 2) As node 4 is a leaf node, decoder backtracks and computes message $\h_{42}$. 3) Decoder continues to backtrack as node 2 has no more children. 4) Node 1 expands node 5 and predicts its label.}
\label{fig:treedecode}
\vspace{-8pt}
\end{figure}

We decode a junction tree $\tree$ from its encoding $\z_\tree$ with a tree structured decoder. The tree is constructed in a top-down fashion by generating one node at a time. As illustrated in Figure~\ref{fig:treedecode}, our tree decoder traverses the entire tree from the root, and generates nodes in their depth-first order. For every visited node, the decoder first makes a \emph{topological prediction}: whether this node has children to be generated. When a new child node is created, we predict its label and recurse this process. Recall that cluster labels represent subgraphs in a molecule. The decoder backtracks when a node has no more children to generate.

At each time step, a node receives information from other nodes in the current tree for making those predictions. The information is propagated through message vectors $\h_{ij}$ when trees are incrementally constructed. Formally, let $\tilde{\calE}=\{(i_1,j_1),\cdots,(i_m,j_m)\}$ be the edges traversed in a depth first traversal over $\tree=(\calV,\calE)$, where $m = 2\len{\calE}$ as each edge is traversed in both directions. The model visits node $i_t$ at time $t$. Let $\tilde{\calE}_t$ be the first $t$ edges in $\tilde{\calE}$. The message $\h_{i_t,j_t}$ is updated through previous messages:
\begin{equation}
\h_{i_t,j_t}=\GRU(\x_{i_t}, \{\h_{k,i_t}\}_{(k,i_t) \in \tilde{\calE}_t, k \neq j_t})
\label{eq:mp}
\end{equation}
where $\GRU$ is the same recurrent unit as in the tree encoder.

\begin{algorithm}[t]
   \caption{Tree decoding at sampling time}
   \label{alg:treedecode}
\renewcommand\algorithmiccomment[1]{\hfill $\triangleright$ {#1}}
\begin{algorithmic}[1]
   \REQUIRE Latent representation $\z_\tree$
   \STATE \textbf{Initialize:} Tree $\widehat{\tree}\leftarrow \emptyset$
   \FUNCTION{SampleTree($i,t$)}
   \STATE Set $\mathcal{X}_i \leftarrow$ all cluster labels that are chemically compatible with node $i$ and its current neighbors.
   \STATE Set $d_t \leftarrow expand$ with probability $p_t$. \hfill $\triangleright$ Eq.(\ref{eq:topo})
   \IF{$d_t=expand$ \AND $\mathcal{X}_i \neq \emptyset$}
   \STATE Create a node $j$ and add it to tree $\widehat{\tree}$.
   \STATE Sample the label of node $j$ from $\mathcal{X}_i$ \hfill $\triangleright$. Eq.(\ref{eq:label})
   \STATE SampleTree($j, t+1$)
   \ENDIF
   \ENDFUNCTION
\end{algorithmic}
\end{algorithm}

\textbf{Topological Prediction } When the model visits node $i_t$, it makes a binary prediction on whether it still has children to be generated. We compute this probability by combining $\z_\tree$, node features $\x_{i_t}$ and inward messages $\h_{k,i_t}$ via a one hidden layer network followed by a sigmoid function:
\begin{equation}
p_t = \sigma(\mbf{u}^d \cdot \tau(\W_1^d \x_{i_t} + \W_2^d \z_\tree + \W_3^d \sum_{(k,i_t) \in \tilde{\calE}_t} \h_{k,i_t})
\label{eq:topo}
\end{equation}
\textbf{Label Prediction } When a child node $j$ is generated from its parent $i$, we predict its node label with 
\begin{equation}
\mbf{q}_j = \softmax(\U^l \tau(\W_1^l \z_\tree + \W_2^l \h_{ij}))
\label{eq:label}
\end{equation}
where $\mbf{q}_j$ is a distribution over label vocabulary $\mathcal{X}$. When $j$ is a root node, its parent $i$ is a virtual node and $\h_{ij} =\mbf{0}$.

\textbf{Learning } The tree decoder aims to maximize the likelihood $p(\tree | \z_\tree)$. Let $\hat{p}_t \in \{0,1\}$ and $\mbf{\hat{q}}_j$ be the ground truth topological and label values, the decoder minimizes the following cross entropy loss:\footnote{The node ordering is not unique as the order within sibling nodes is ambiguous. In this paper we train our model with one ordering and leave this issue for future work.}
\begin{equation}
\loss_c(\tree) = \sum\nolimits_t \loss^d(p_t,\hat{p}_t) + \sum\nolimits_j \loss^l(\mbf{q}_j,\mbf{\hat{q}}_j)
\end{equation}
Similar to sequence generation, during training we perform \emph{teacher forcing}: after topological and label prediction at each step, we replace them with their ground truth so that the model makes predictions given correct histories.

\textbf{Decoding \& Feasibility Check } Algorithm~\ref{alg:treedecode} shows how a tree is sampled from $\z_\tree$. The tree is constructed recursively guided by topological predictions without any external guidance used in training. To ensure the sampled tree could be realized into a valid molecule, we define set $\mathcal{X}_i$ to be cluster labels that are chemically compatible with node $i$ and its current neighbors. When a child node $j$ is generated from node $i$, we sample its label from $\mathcal{X}_i$ with a renormalized distribution $\mbf{q}_j$ over $\mathcal{X}_i$ by masking out invalid labels.
\subsection{Graph Decoder}
\label{sec:gdecode}
The final step of our model is to reproduce a molecular graph $G$ that underlies the predicted junction tree $\widehat{\tree}=(\widehat{\calV},\widehat{\calE})$. Note that this step is not deterministic since there are potentially many molecules that correspond to the same junction tree. The underlying degree of freedom pertains to how neighboring clusters $C_i$ and $C_j$ are attached to each other as subgraphs. Our goal here is to assemble the subgraphs (nodes in the tree) together into the correct molecular graph. 

\begin{figure}[t]
\centering
\includegraphics[width=0.48\textwidth]{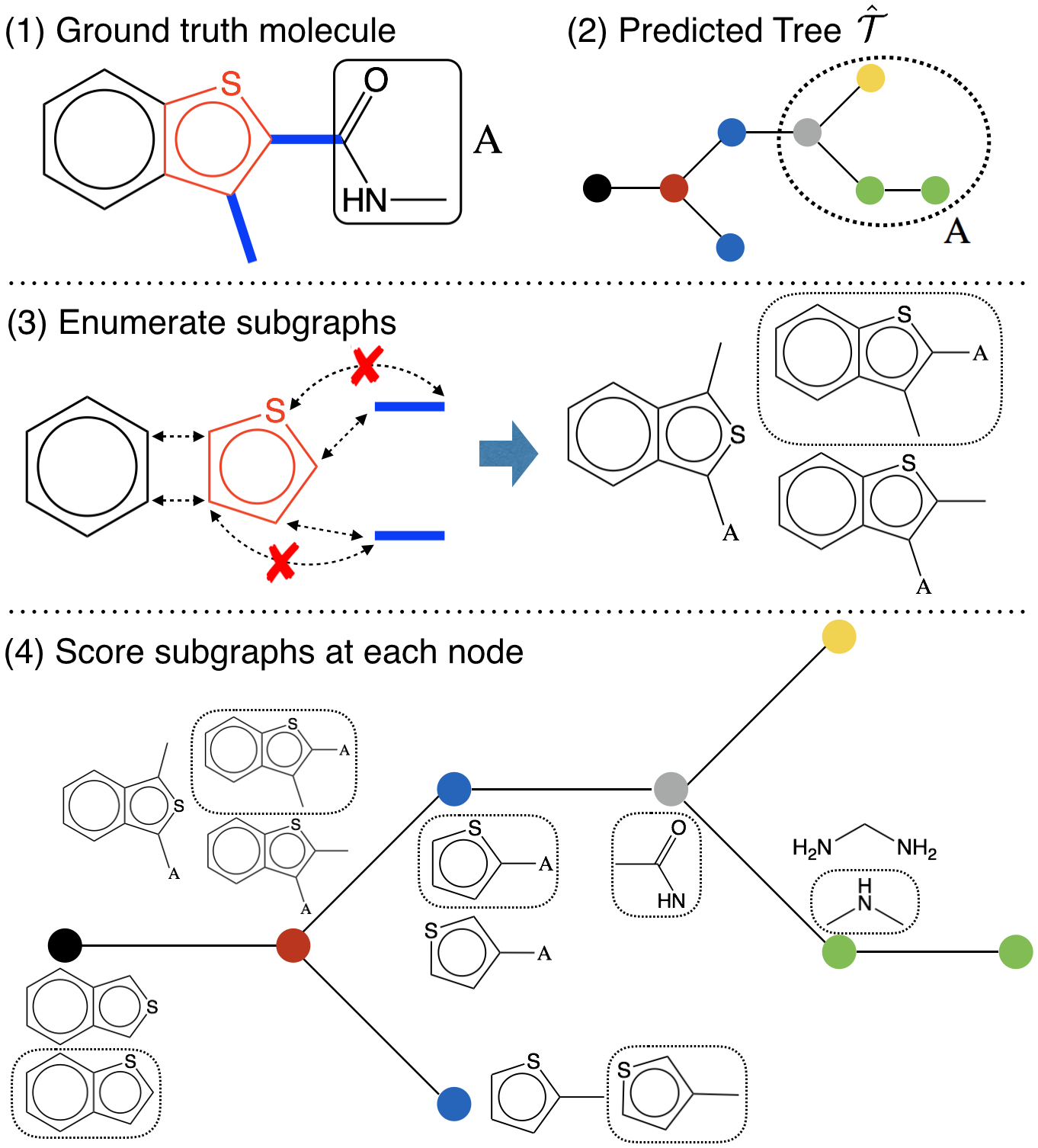}
\vspace{-18pt}
\caption{Decode a molecule from a junction tree. 1) Ground truth molecule $G$. 2) Predicted junction tree $\widehat{\tree}$. 3) We enumerate different combinations between red cluster $C$ and its neighbors. Crossed arrows indicate combinations that lead to chemically infeasible molecules. Note that if we discard tree structure during enumeration (i.e., ignoring subtree A), the last two candidates will collapse into the same molecule. 4) Rank subgraphs at each node. The final graph is decoded by putting together all the predicted subgraphs (dashed box).}
\label{fig:assm}
\vspace{-8pt}
\end{figure}

Let $\G(\tree)$ be the set of graphs whose junction tree is $\tree$. Decoding graph $\hat{G}$ from $\widehat{\tree}=(\widehat{\calV},\widehat{\calE})$ is a structured prediction:
\begin{equation}
\hat{G} = \arg\max_{G'\in \G(\widehat{\tree})} f^a(G')
\end{equation}
where $f^a$ is a scoring function over candidate graphs. We only consider scoring functions that decompose across the clusters and their neighbors. In other words, each term in the scoring function depends only on how a cluster $C_i$ is attached to its neighboring clusters  $C_j$, $j\in N_{\widehat{\tree}}(i)$ in the tree $\widehat{\tree}$. The problem of finding the highest scoring graph $\hat{G}$ -- the assembly task -- could be cast as a graphical model inference task in a model induced by the junction tree. However, for efficiency reasons, we will assemble the molecular graph one neighborhood at a time, following the order in which the tree itself was decoded. In other words, we start by sampling the assembly of the root and its neighbors according to their scores. Then we proceed to assemble the neighbors and their associated clusters (removing the degrees of freedom set by the root assembly), and so on. 

It remains to be specified how each neighborhood realization is scored. Let $G_i$ be the subgraph resulting from a particular merging of cluster $C_i$ in the tree with its neighbors $C_j$, $j\in N_{\widehat{\tree}}(i)$. We score $G_i$ as a candidate subgraph by first deriving a vector representation $\h_{G_i}$ and then using $f^a_i(G_i) = \h_{G_i}\cdot \z_G$ as the subgraph score. To this end, let $u,v$ specify atoms in the candidate subgraph $G_i$ and let $\alpha_v=i$ if $v\in C_i$ and $\alpha_v = j$ if $v\in C_j\setminus C_i$. The indices $\alpha_v$ are used to mark the position of the atoms in the junction tree, and to retrieve messages $\widehat{\m}_{i,j}$ summarizing the subtree under $i$ along the edge $(i,j)$ obtained by running the tree encoding algorithm. The neural messages pertaining to the atoms and bonds in subgraph $G_i$ are obtained and aggregated into $\h_{G_i}$, similarly to the encoding step, but with different (learned) parameters:
\begin{eqnarray}
\bmu_{uv}^{(t)} &=& \tau(\W_1^a \x_u + \W_2^a \x_{uv} + \W_3^a \widetilde{\bmu}_{uv}^{(t-1)}) \\
\widetilde{\bmu}_{uv}^{(t-1)} &=&
\begin{cases}
\sum_{w \in N(u) \backslash v} \bmu_{wu}^{(t-1)} & \alpha_u =\alpha_v \\
\widehat{\m}_{\alpha_u,\alpha_v} + \sum_{w \in N(u) \backslash v} \bmu_{wu}^{(t-1)} & \alpha_u \neq \alpha_v
\end{cases} \nonumber
\end{eqnarray}
The major difference from Eq.~(\ref{eq:mpn}) is that we augment the model with tree messages $\widehat{\m}_{\alpha_u,\alpha_v}$ derived by running the tree encoder over the predicted tree $\widehat{\tree}$. $\widehat{\m}_{\alpha_u,\alpha_v}$ provides a tree dependent positional context for bond $(u,v)$ (illustrated as subtree A in Figure~\ref{fig:assm}).

\textbf{Learning } The graph decoder parameters are learned to maximize the log-likelihood of predicting correct subgraphs $G_i$ of the ground true graph $G$ at each tree node: 
\begin{equation}
\loss_g(G) = \sum_i \left[f^a(G_i) - \log \sum_{G_i' \in {\cal G}_i} \exp(f^a(G_i'))\right]
\end{equation}
where ${\cal G}_i$ is the set of possible candidate subgraphs at tree node $i$. During training, we again apply teacher forcing, i.e. we feed the graph decoder with ground truth trees as input. 

\textbf{Complexity} By our tree decomposition, any two clusters share at most two atoms, so we only need to merge at most two atoms or one bond. By pruning chemically invalid subgraphs and merging isomorphic graphs, $|{\cal G}_i|\approx 4$ on average when tested on a standard ZINC drug dataset. The computational complexity of JT-VAE is therefore linear in the number of clusters, scaling nicely to large graphs.
\section{Experiments}

\begin{figure*}[t]
\vspace{-5pt}
\centering
\begin{subfigure}[t]{0.45\textwidth}
\centering
\includegraphics[width=\textwidth]{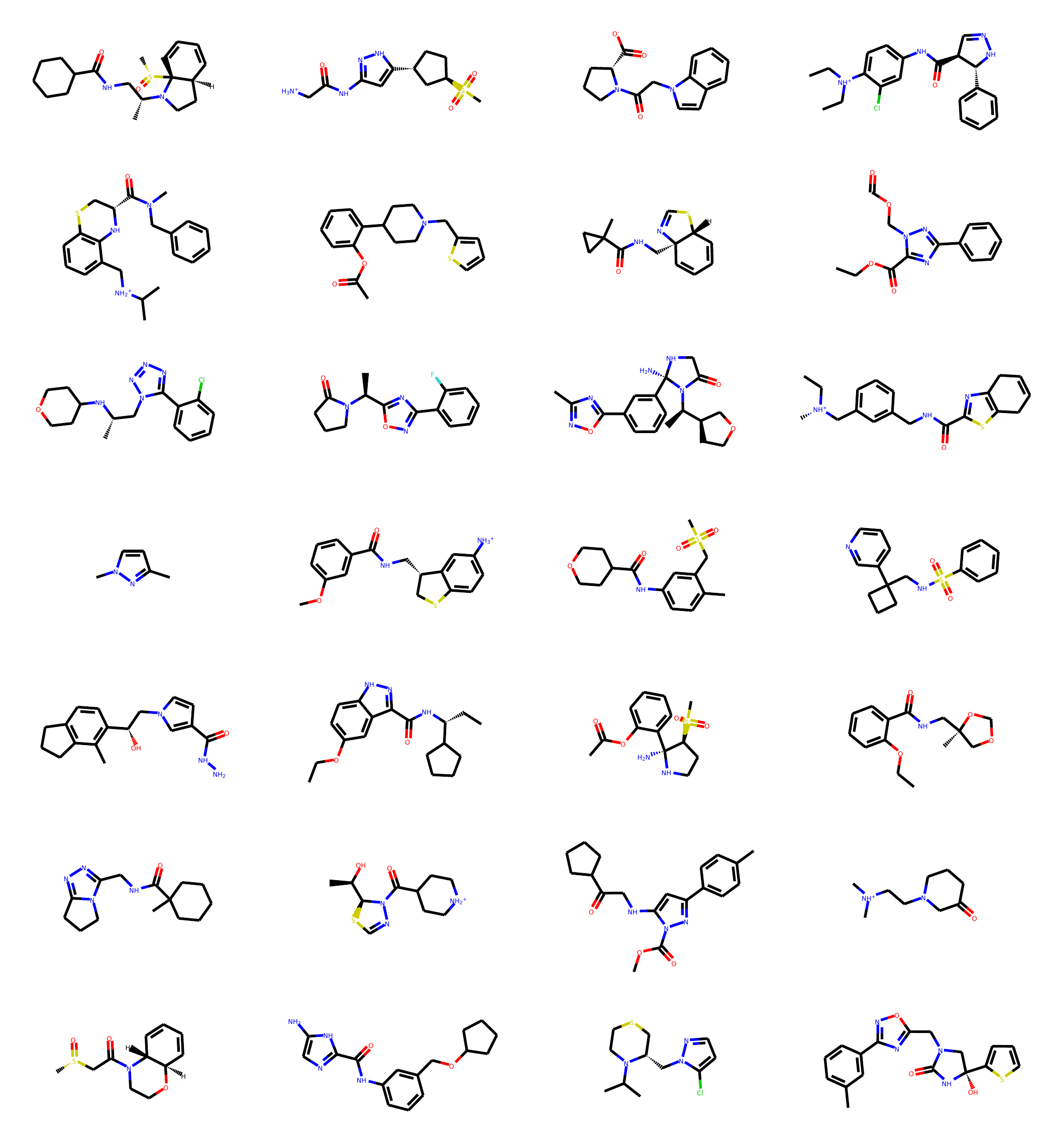}
\end{subfigure}
~
\begin{subfigure}[t]{0.49\textwidth}
\centering
\includegraphics[width=\textwidth]{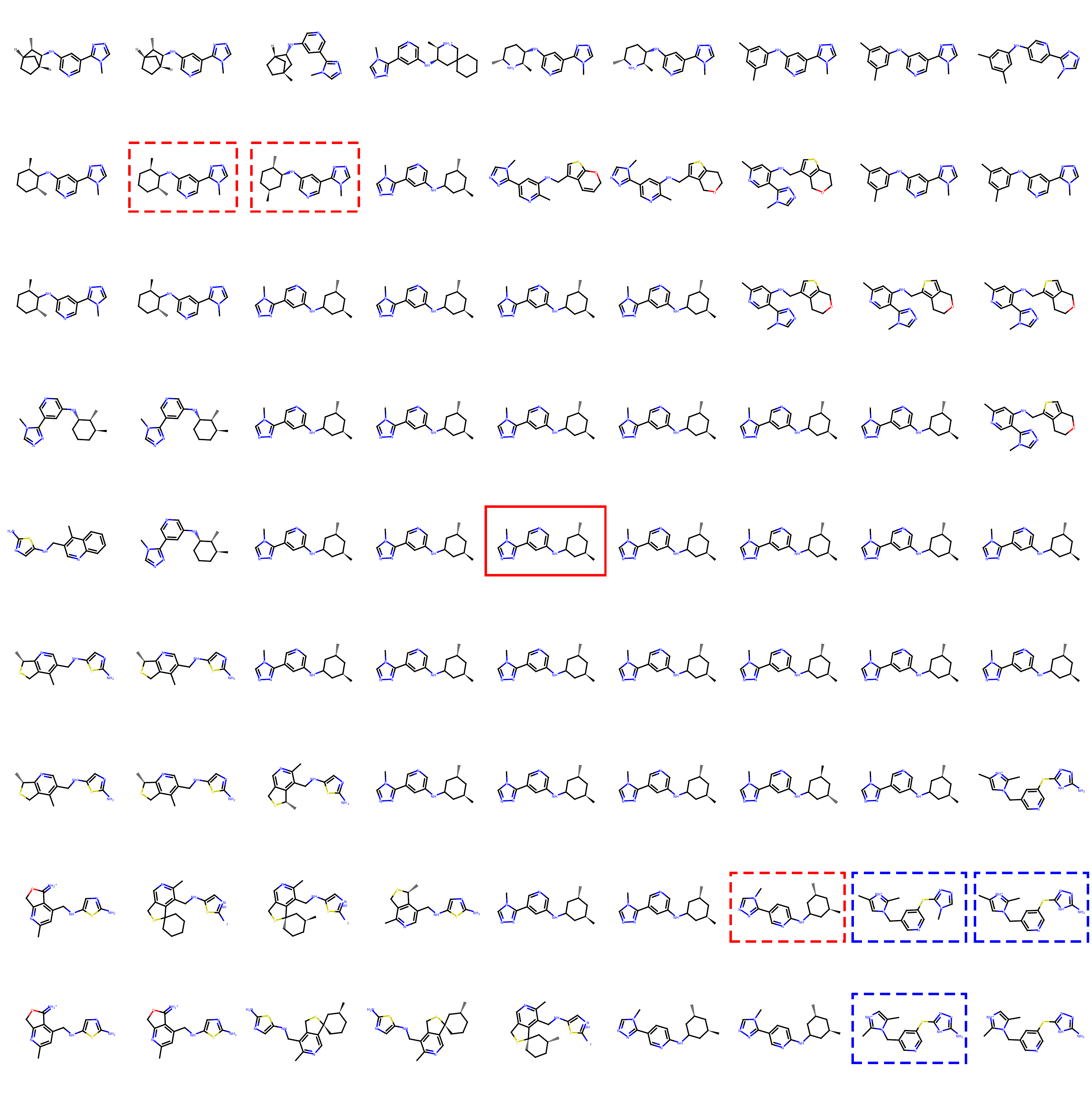}
\end{subfigure}
\vspace{-10pt}
\caption{\textbf{Left}: Random molecules sampled from prior distribution $\prior$. \textbf{Right}: Visualization of the local neighborhood of a molecule in the center. Three molecules highlighted in red dashed box have the same tree structure as the center molecule, but with different graph structure as their clusters are combined differently. The same phenomenon emerges in another group of molecules (blue dashed box).}
\label{fig:vis}
\vspace{-10pt}
\end{figure*}

Our evaluation efforts measure various aspects of molecular generation. The first two evaluations
follow previously proposed tasks~\cite{kusner2017grammar}. We also introduce a third task --- constrained molecule optimization. 
\vspace{-6pt}
\begin{itemize}[leftmargin=*] \itemsep=-1pt
\item \textbf{Molecule reconstruction and validity} We test the VAE models on the task of reconstructing input molecules from their latent representations, and decoding valid molecules when sampling from prior distribution. (Section~\ref{sec:molgen})

\item \textbf{Bayesian optimization} Moving beyond generating valid molecules, we test how the model can produce novel molecules with desired properties. To this end, we perform Bayesian optimization in the latent space to search molecules with specified properties. (Section~\ref{sec:bo})

\item \textbf{Constrained molecule optimization} The task is to modify given molecules to improve specified properties, while constraining the degree of deviation from the original molecule. This is a more realistic scenario in drug discovery, where development of new drugs usually starts with known molecules such as existing drugs~\cite{besnard2012automated}. Since it is a new task, we cannot compare to any existing baselines. (Section~\ref{sec:conopt})
\end{itemize}
\vspace{-5pt}
Below we describe the data, baselines and model configuration that are shared across the tasks. Additional setup details are provided in the task-specific sections.

\textbf{Data } We use the ZINC molecule dataset from \citet{kusner2017grammar} for our experiments, with the same training/testing split. It contains about 250K drug molecules extracted from the ZINC database~\citep{sterling2015zinc}. We follow the same train/test split as in \citet{kusner2017grammar}.

\textbf{Baselines } We compare our approach with SMILES-based baselines: 1) Character VAE (CVAE)~\cite{gomez2016automatic} which generates SMILES strings character by character; 2) Grammar VAE (GVAE)~\cite{kusner2017grammar} that generates SMILES following syntactic constraints given by a context-free grammar; 3) Syntax-directed VAE (SD-VAE)~\cite{dai2018syntax-directed} that incorporates both syntactic and semantic constraints of SMILES via attribute grammar.
For molecule generation task, we also compare with GraphVAE~\cite{simonovsky2018graphvae} that directly generates atom labels and adjacency matrices of graphs, as well as an LSTM-based autoregressive model that generates molecular graphs atom by atom~\citep{li2018learning}.

\textbf{Model Configuration } To be comparable with the above baselines, we set the latent space dimension as 56, i.e., the tree and graph representation $\h_\tree$ and $\h_G$ have 28 dimensions each. Full training details and model configurations are provided in the appendix.

\subsection{Molecule Reconstruction and Validity} 
\label{sec:molgen}

\textbf{Setup } The first task is to reconstruct and sample molecules from latent space. Since both encoding and decoding process are stochastic, we estimate reconstruction accuracy by Monte Carlo method used in \cite{kusner2017grammar}: Each molecule is encoded 10 times and each encoding is decoded 10 times. We report the portion of the 100 decoded molecules that are identical to the input molecule. 

To compute validity, we sample 1000 latent vectors from the prior distribution $\prior$, and decode each of these vectors 100 times. We report the percentage of decoded molecules that are chemically valid (checked by RDKit). For ablation study, we also report the validity of our model without validity check in decoding phase.

\textbf{Results} Table~\ref{tab:basic} shows that JT-VAE outperforms previous models in molecule reconstruction, and \textbf{always} produces valid molecules when sampled from prior distribution. In contrast, the atom-by-atom based generation only achieves 89.2\% validity as it needs to go through invalid intermediate states (Figure~\ref{fig:structgen}). Our model bypasses this issue by utilizing valid substructures as building blocks. As shown in Figure~\ref{fig:vis}, the sampled molecules have non-trivial structures such as simple chains. We further sampled 5000 molecules from prior and found they are \emph{all distinct} from the training set. Thus our model is not a simple memorization.

\begin{table}[tb]
\vspace{-7pt}
\centering
\caption{Reconstruction accuracy and prior validity results. Baseline results are copied from \citet{kusner2017grammar,dai2018syntax-directed,simonovsky2018graphvae,li2018learning}.}
\vspace{2pt}

\newcolumntype{C}{>{\centering\arraybackslash}X}
\begin{tabularx}{0.48\textwidth}{lCC}
\hline
\textbf{Method} & \textbf{Reconstruction} & \textbf{Validity} \Tstrut\Bstrut \\
\hline
CVAE & 44.6\% & 0.7\% \Tstrut\Bstrut \\
GVAE & 53.7\% & 7.2\% \Tstrut\Bstrut \\
SD-VAE & 76.2\% & 43.5\% \Tstrut\Bstrut \\
GraphVAE & - & 13.5\% \Tstrut\Bstrut \\
Atom-by-Atom LSTM & - & 89.2\% \Tstrut\Bstrut \\
\hline
JT-VAE & \textbf{76.7\%} & \textbf{100.0\%} \Tstrut\Bstrut \\
\hline
\end{tabularx}
\label{tab:basic}
\vspace{-12pt}
\end{table}

\textbf{Analysis } We qualitatively examine the latent space of JT-VAE by visualizing the neighborhood of molecules. Given a molecule, we follow the method in \citet{kusner2017grammar} to construct a grid visualization of its neighborhood. 
Figure~\ref{fig:vis} shows the local neighborhood of the same molecule visualized in \citet{dai2018syntax-directed}. In comparison, our neighborhood does not contain molecules with huge rings (with more than 7 atoms), which rarely occur in the dataset. We also highlight two groups of closely resembling molecules that have identical tree structures but vary only in how clusters are attached together. This demonstrates the smoothness of learned molecular embeddings.

\subsection{Bayesian Optimization} 
\label{sec:bo}

\textbf{Setup } The second task is to produce novel molecules with desired properties. Following~\cite{kusner2017grammar}, our target chemical property $y(\cdot)$ is octanol-water partition coefficients (logP) penalized by the synthetic accessibility (SA) score and number of long cycles.\footnote{$y(m) = logP(m) - SA(m) - cycle(m)$ where $cycle(m)$ counts the number of rings that have more than six atoms.}
To perform Bayesian optimization (BO), we first train a VAE and associate each molecule with a latent vector, given by the mean of the variational encoding distribution. After the VAE is learned, we train a sparse Gaussian process (SGP) to predict $y(m)$ given its latent representation. Then we perform five iterations of batched BO using the expected improvement heuristic.

For comparison, we report 1) the predictive performance of SGP trained on latent encodings learned by different VAEs, measured by log-likelihood (LL) and root mean square error (RMSE) with 10-fold cross validation. 2) The top-3 molecules found by BO under different models.

\begin{figure}[t]
\centering
\includegraphics[width=0.48\textwidth]{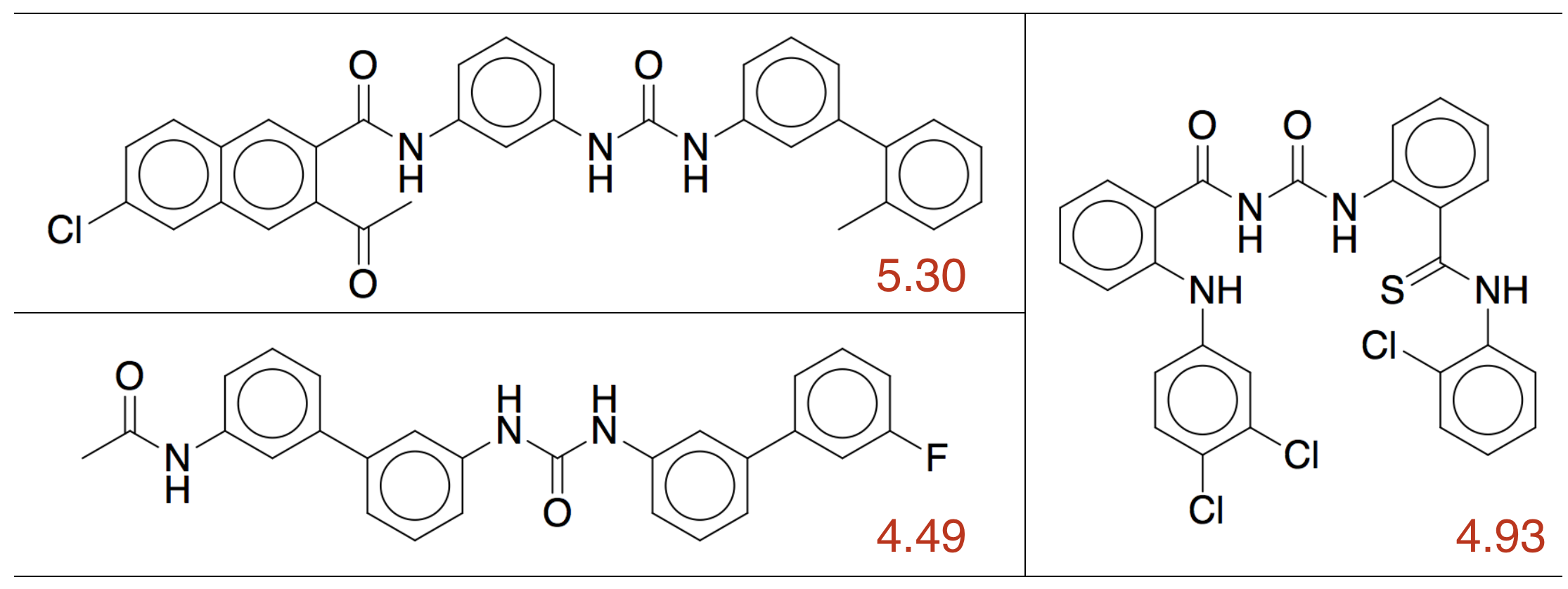}
\vspace{-18pt}
\caption{Best three molecules and their property scores found by JT-VAE using Bayesian optimization.}
\label{fig:bo}
\vspace{-10pt}
\end{figure}

\textbf{Results } As shown in Table~\ref{tab:bo}, JT-VAE finds molecules with significantly better scores than previous methods. Figure~\ref{fig:bo} lists the top-3 best molecules found by JT-VAE. In fact, JT-VAE finds over 50 molecules with scores over 3.50 (the second best molecule proposed by SD-VAE). Moreover, the SGP yields better predictive performance when trained on JT-VAE embeddings (Table~\ref{tab:sgp}).

\begin{table}[t]
\centering
\vspace{-7pt}
\caption{Best molecule property scores found by each method. Baseline results are from \citet{kusner2017grammar,dai2018syntax-directed}.}
\vspace{5pt}

\newcolumntype{C}{>{\centering\arraybackslash}X}
\begin{tabularx}{0.45\textwidth}{CCCC}
\hline
\textbf{Method} & \textbf{1st} & \textbf{2nd} & \textbf{3rd} \Tstrut\Bstrut \\
\hline
CVAE & 1.98 & 1.42 & 1.19 \Tstrut\Bstrut \\
GVAE & 2.94 & 2.89 & 2.80 \Tstrut\Bstrut \\
SD-VAE & 4.04 & 3.50 & 2.96 \Tstrut\Bstrut \\
\hline
JT-VAE & \textbf{5.30} & \textbf{4.93} & \textbf{4.49} \Tstrut\Bstrut \\
\hline
\end{tabularx}
\label{tab:bo}
\end{table}

\begin{table}[t]
\centering
\vspace{-7pt}
\caption{Predictive performance of sparse Gaussian Processes trained on different VAEs. Baseline results are copied from \citet{kusner2017grammar} and \citet{dai2018syntax-directed}.}
\vspace{5pt}
\begin{tabular}{ccc}
\hline
\textbf{Method} & \textbf{LL} & \textbf{RMSE} \Tstrut\Bstrut \\
\hline
CVAE & $-1.812 \pm 0.004$ & $1.504 \pm 0.006$ \Tstrut\Bstrut \\
GVAE & $-1.739 \pm 0.004$ & $1.404 \pm 0.006$ \Tstrut\Bstrut \\
SD-VAE & $-1.697 \pm 0.015$ & $1.366 \pm 0.023$ \Tstrut\Bstrut \\
\hline
JT-VAE & $\mbf{-1.658 \pm 0.023}$ & $\mbf{1.290 \pm 0.026}$ \Tstrut\Bstrut \\
\hline
\end{tabular}
\label{tab:sgp}
\end{table}

\begin{table}[t]
\centering
\vspace{-10pt}
\caption{Constrained optimization result of JT-VAE: mean and standard deviation of property improvement, molecular similarity and success rate under constraints $sim(m,m') \geq \delta$ with varied $\delta$.}
\vspace{5pt}
\begin{tabular}{cccc}
\hline
$\delta$ & \textbf{Improvement} & \textbf{Similarity} & \textbf{Success} \Tstrut\Bstrut \\
\hline
0.0 & $1.91 \pm 2.04$ & $0.28 \pm 0.15$ & 97.5\% \Tstrut\Bstrut \\
0.2 & $1.68 \pm 1.85$ & $0.33 \pm 0.13$ & 97.1\% \Tstrut\Bstrut \\
0.4 & $0.84 \pm 1.45$ & $0.51 \pm 0.10$ & 83.6\% \Tstrut\Bstrut \\
0.6 & $0.21 \pm 0.71$ & $0.69 \pm 0.06$ & 46.4\% \Tstrut\Bstrut \\
\hline
\end{tabular}
\label{tab:copt}
\vspace{-10pt}
\end{table}

\begin{figure}[t]
\centering
\includegraphics[width=0.45\textwidth]{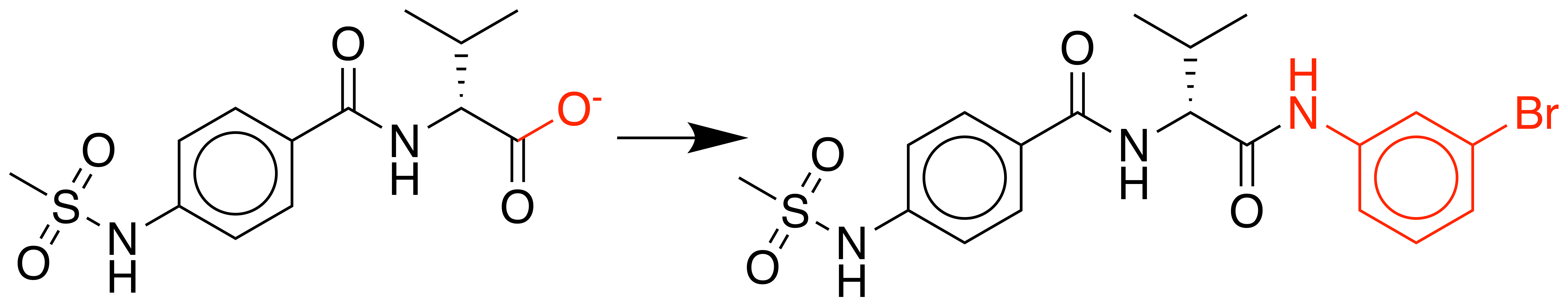}
\vspace{-8pt}
\caption{A molecule modification that yields an improvement of 4.0 with molecular similarity 0.617 (modified part is in red).}
\label{fig:copt}
\vspace{-15pt}
\end{figure}

\subsection{Constrained Optimization}
\label{sec:conopt}

\textbf{Setup } The third task is to perform molecule optimization in a constrained scenario. Given a molecule $m$, the task is to find a different molecule $m'$ that has the highest property value with the molecular similarity $sim(m,m')\geq \delta$ for some threshold $\delta$. We use Tanimoto similarity with Morgan fingerprint~\cite{rogers2010extended} as the similarity metric, and penalized logP coefficient as our target chemical property. 
For this task, we jointly train a property predictor $F$ (parameterized by a feed-forward network) with JT-VAE to predict $y(m)$ from the latent embedding of $m$.
To optimize a molecule $m$, we start from its latent representation, and apply gradient ascent in the latent space to improve the predicted score $F(\cdot)$, similar to \cite{mueller2017sequence}. After applying $K=80$ gradient steps, $K$ molecules are decoded from resulting latent trajectories, and we report the molecule with the highest $F(\cdot)$ that satisfies the similarity constraint. A modification succeeds if one of the decoded molecules satisfies the constraint and is distinct from the original.

To provide the greatest challenge, we selected 800 molecules with the \emph{lowest} property score $y(\cdot)$ from the test set. We report the success rate (how often a modification succeeds), and among success cases the average improvement $y(m')-y(m)$ and molecular similarity $sim(m,m')$ between the original and modified molecules $m$ and $m'$.

\textbf{Results } Our results are summarized in Table~\ref{tab:copt}. The unconstrained scenario ($\delta=0$) has the best average improvement, but often proposes dissimilar molecules.
When we tighten the constraint to $\delta=0.4$, about 80\% of the time our model finds similar molecules, with an average improvement $0.84$. This also demonstrates the smoothness of the learned latent space. Figure~\ref{fig:copt} illustrates an effective modification resulting in a similar molecule with great improvement. 
\section{Related Work}
\textbf{Molecule Generation }
Previous work on molecule generation mostly operates on SMILES strings. \citet{gomez2016automatic,segler2017generating} built generative models of SMILES strings with recurrent decoders. Unfortunately, these models could generate invalid SMILES that do not result in any molecules. 
To remedy this issue, \citet{kusner2017grammar,dai2018syntax-directed} complemented the decoder with syntactic and semantic constraints of SMILES by context free and attribute grammars, but these grammars do not fully capture chemical validity. Other techniques such as active learning~\cite{janz2017actively} and reinforcement learning~\cite{guimaraes2017objective} encourage the model to generate valid SMILES through additional training signal. 
Very recently, \citet{simonovsky2018graphvae} proposed to generate molecular graphs by predicting their adjacency matrices, and \citet{li2018learning} generated molecules node by node. In comparison, our method enforces chemical validity and is more efficient due to the coarse-to-fine generation.

\textbf{Graph-structured Encoders }
The neural network formulation on graphs was first proposed by \citet{gori2005graph,scarselli2009graph}, and  later enhanced by \citet{li2015gated} with gated recurrent units. For recurrent architectures over graphs, \citet{lei2017deriving} designed Weisfeiler-Lehman kernel network inspired by  graph kernels. \citet{dai2016discriminative} considered a different architecture where graphs were viewed as latent variable graphical models, and derived their model from message passing algorithms. Our tree and graph encoder are closely related to this graphical model perspective, and to neural message passing networks~\cite{gilmer2017neural}.
For convolutional architectures, \citet{duvenaud2015convolutional} introduced a convolution-like propagation on molecular graphs, which was generalized to other domains by \citet{niepert2016learning}. \citet{bruna2013spectral,henaff2015deep} developed graph convolution in spectral domain via graph Laplacian. For applications, graph neural networks are used in semi-supervised classification~\cite{kipf2016semi}, computer vision~\cite{monti2016geometric}, and chemical domains~\cite{kearnes2016molecular,schutt2017schnet,jin2017predicting}.

\textbf{Tree-structured Models } Our tree encoder is related to recursive neural networks and tree-LSTM \cite{socher2013recursive,tai2015improved,zhu2015long}. These models encode tree structures where nodes in the tree are bottom-up transformed into vector representations. In contrast, our model propagates information both bottom-up and top-down.

On the decoding side, tree generation naturally arises in natural language parsing~\citep{dyer2016recurrent,kiperwasser2016easy}.
Different from our approach, natural language parsers have access to input words and only predict the topology of the tree. 
For general purpose tree generation, \citet{vinyals2015grammar,aharoni2017towards} applied recurrent networks to generate linearized version of trees, but their architectures were entirely sequence-based. 
\citet{dong2016language,alvarez2016tree} proposed tree-based architectures that construct trees top-down from the root. Our model is most closely related to \citet{alvarez2016tree} that disentangles topological prediction from label prediction, but we generate nodes in a depth-first order and have additional steps that propagate information bottom-up. This forward-backward propagation also appears in \citet{parisotto2016neuro}, but their model is node based whereas ours is based on message passing.
\section{Conclusion}
In this paper we present a junction tree variational autoencoder for generating molecular graphs. Our method significantly outperforms previous work in molecule generation and optimization. For future work, we attempt to generalize our method for general low-treewidth graphs.

\section*{Acknowledgement}
We thank Jonas Mueller, Chengtao Li, Tao Lei and MIT NLP Group for their helpful comments. This work was supported by the DARPA Make-It program under contract ARO W911NF-16-2-0023.
\bibliography{main}
\bibliographystyle{icml2018}
\onecolumn
\appendix
\icmltitle{Supplementary Material}
\section{Tree Decomposition}
Algorithm~\ref{alg:treedecomp} presents our tree decomposition of molecules. $V_1$ and $V_2$ contain non-ring bonds and simple rings respectively. Simple rings are extracted via RDKit's \texttt{GetSymmSSSR} function. We then merge rings that share three or more atoms as they form bridged compounds. 
We note that the junction tree of a molecule is not unique when its cluster graph contains cycles. This introduces additional uncertainty for our probabilistic modeling. To reduce such variation, for any of the three (or more) intersecting bonds, we add their intersecting atom as a cluster and remove the cycle connecting them in the cluster graph. 
Finally, we construct a junction tree as the maximum spanning tree of a cluster graph $(\calV,\calE)$. Note that we assign an large weight over edges involving clusters in $V_0$ to ensure no edges in any cycles will be selected into the junction tree.

\begin{algorithm}[h]
\caption{Tree decomposition of molecule $G=(V,E)$}
\label{alg:treedecomp}
\renewcommand\algorithmiccomment[1]{\hfill {\small \# \textit{#1}}}
\begin{algorithmic}
    \STATE $V_1 \leftarrow$ the set of bonds $(u,v)\in E$ that do not belong to any rings.
    \STATE $V_2 \leftarrow$ the set of simple rings of $G$.
    \FOR{$r_1,r_2 $ {\bfseries in} $V_2$} 
    \STATE Merge rings $r_1,r_2$ into one ring if they share more than two atoms (bridged rings).
    \ENDFOR
    \STATE $V_0 \leftarrow$ atoms being the intersection of three or more clusters in $V_1 \cup V_2$.
    \STATE $\calV \leftarrow V_0 \cup V_1 \cup V_2$
    \STATE $\calE \leftarrow \{ (i,j,c) \in \calV \times \calV \times \real \;|\; \len{i \cap j} > 0\}$. Set $c=\infty$ if $i \in V_0$ or $j \in V_0$, and $c=1$ otherwise.
    \STATE \textbf{Return} The maximum spanning tree over cluster graph $(\calV,\calE)$.
\end{algorithmic}
\end{algorithm}

\begin{figure}[h]
\centering
\vspace{-5pt}
\includegraphics[width=0.9\textwidth]{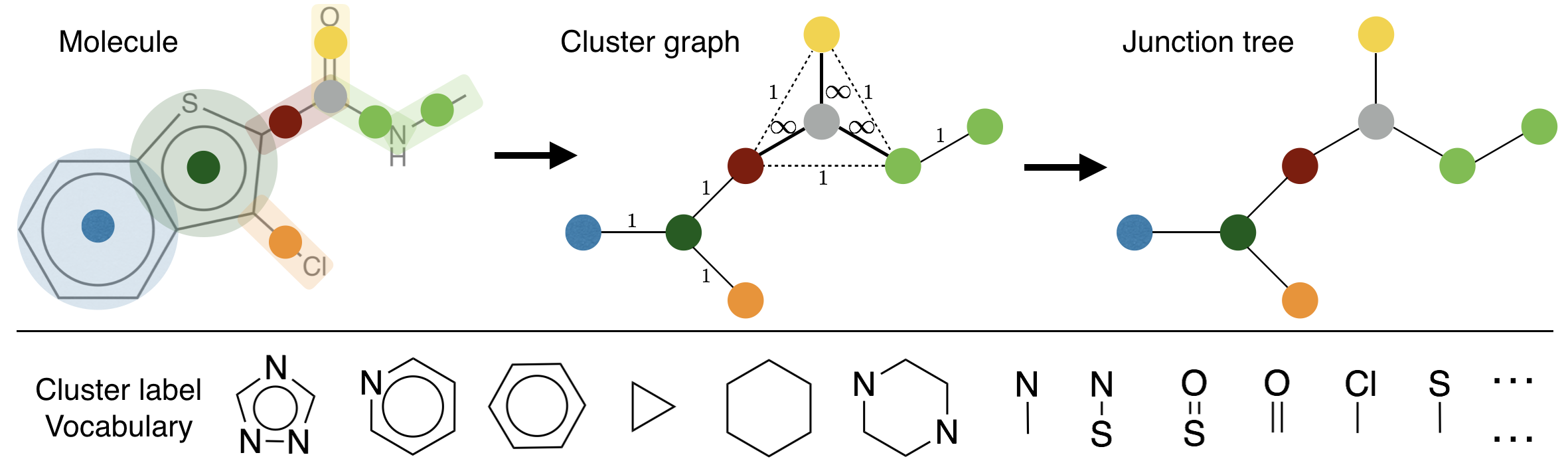}
\vspace{-5pt}
\caption{Illustration of tree decomposition and sample of cluster label vocabulary.}
\vspace{-5pt}
\end{figure}

\section{Stereochemistry}
Though usually presented as two-dimensional graphs, molecules are three-dimensional objects, i.e. molecules are defined not only by its atom types and bond connections, but also the spatial configuration between atoms (chiral atoms and cis-trans isomerism). \emph{Stereoisomers} are molecules that have the same 2D structure, but differ in the 3D orientations of their atoms in space. We note that stereochemical feasibility could not be simply encoded as context free or attribute grammars.

Empirically, we found it more efficient to predict the stereochemical configuration separately from the molecule generation. Specifically, the JT-VAE first generates the 2D structure of a molecule $m$, following the same procedure described in section~\ref{sec:jtnn}. Then we generate all its stereoisomers $\mathcal{S}_m$ using RDKit's \texttt{EnumerateStereoisomers} function, which identifies atoms that could be chiral. For each isomer $m'\in \mathcal{S}_m$, we encode its graph representation $\h_{m'}$ with the graph encoder and compute their cosine similarity $f^s(m') = \cos(\h_{m'},\z_m)$ (note that $\z_m$ is stochastic). We reconstruct the final 3D structure by picking the stereoisomer $\widehat{m} = \arg\max_{m'} f^s(m')$. Since on average only few atoms could have stereochemical variations, this post ranking process is very efficient. Combining this with tree and graph generation, the molecule reconstruction loss $\loss$ becomes
\begin{equation}
\loss = \loss_c + \loss_g + \loss_s; \qquad \loss_s = f^s(m) - \log \sum_{m' \in \mathcal{S}_m} \exp(f^s(m'))
\end{equation}

\section{Training Details}
By applying tree decomposition over 240K molecules in ZINC dataset, we collected our vocabulary set $\mathcal{X}$ of size $\len{\mathcal{X}}=780$.
The hidden state dimension is 450 for all modules in JT-VAE and the latent bottleneck dimension is 56. For the graph encoder, the initial atom features include its atom type, degree, its formal charge and its chiral configuration. Bond feature is a concatenation of its bond type, whether the bond is in a ring, and its cis-trans configuration. For our tree encoder, we represent each cluster with a neural embedding vector, similar to word embedding for words. The tree and graph decoder use the same feature setting as encoders. The graph encoder and decoder runs three iterations of neural message passing. For fair comparison to SMILES based method, we minimized feature engineering. 
We use PyTorch to implement all neural components and RDKit to process molecules.

\section{More Experimental Results}
\begin{figure}[t]
\centering
\includegraphics[width=\textwidth]{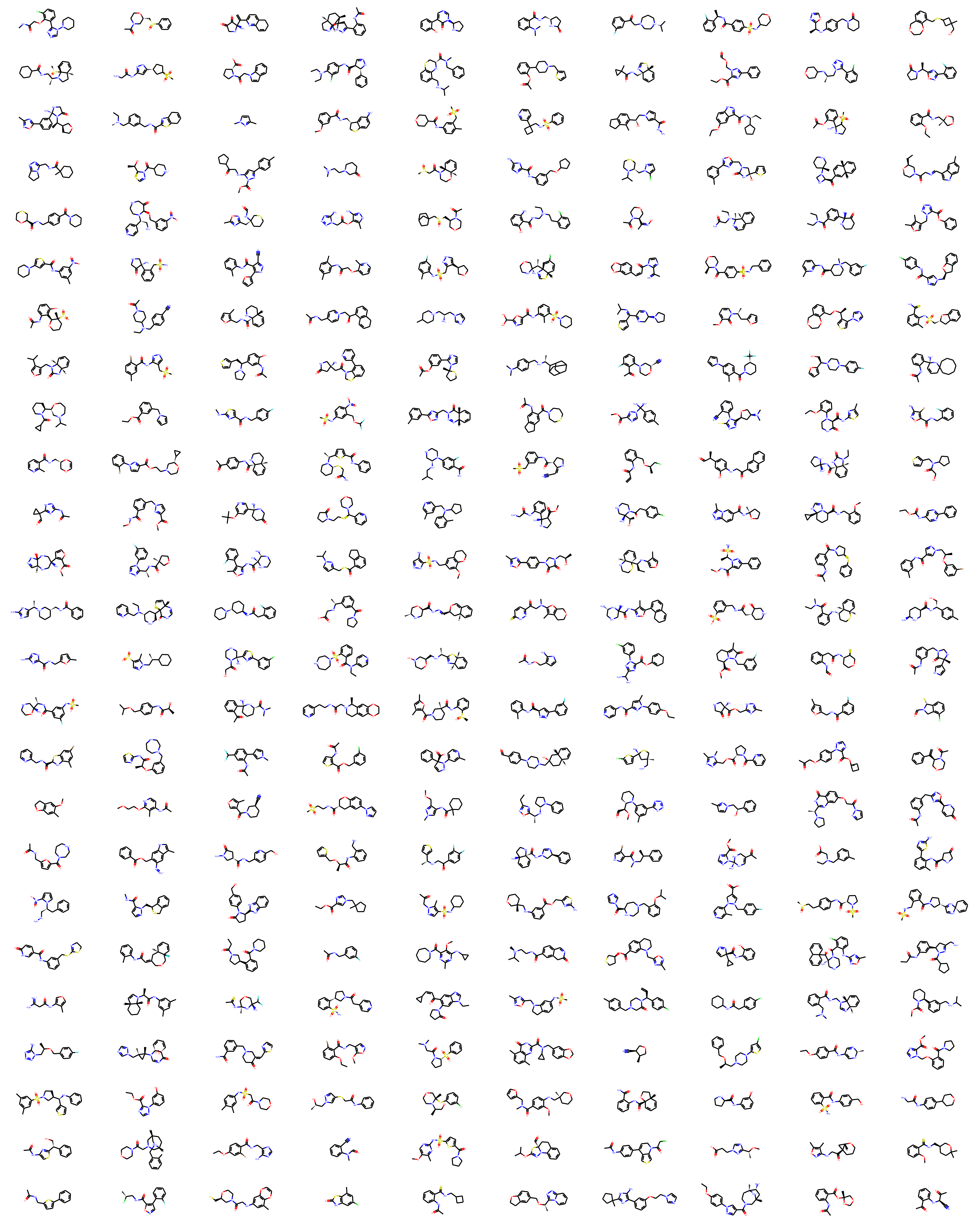}
\caption{250 molecules sampled from prior distribution $\prior$.}
\label{fig:priormols}
\end{figure}

\textbf{Sampled Molecules } Note that a degenerate model could also achieve 100\% prior validity by keep generating simple structures like chains. To prove that our model does not converge to such trivial solutions, we randomly sample and plot 250 molecules from prior distribution $\prior$. As shown in Figure~\ref{fig:priormols}, our sampled molecules present rich variety and structural complexity. This demonstrates the soundness of the prior validity improvement of our model.

\begin{figure}[t]
\centering
\includegraphics[width=\textwidth]{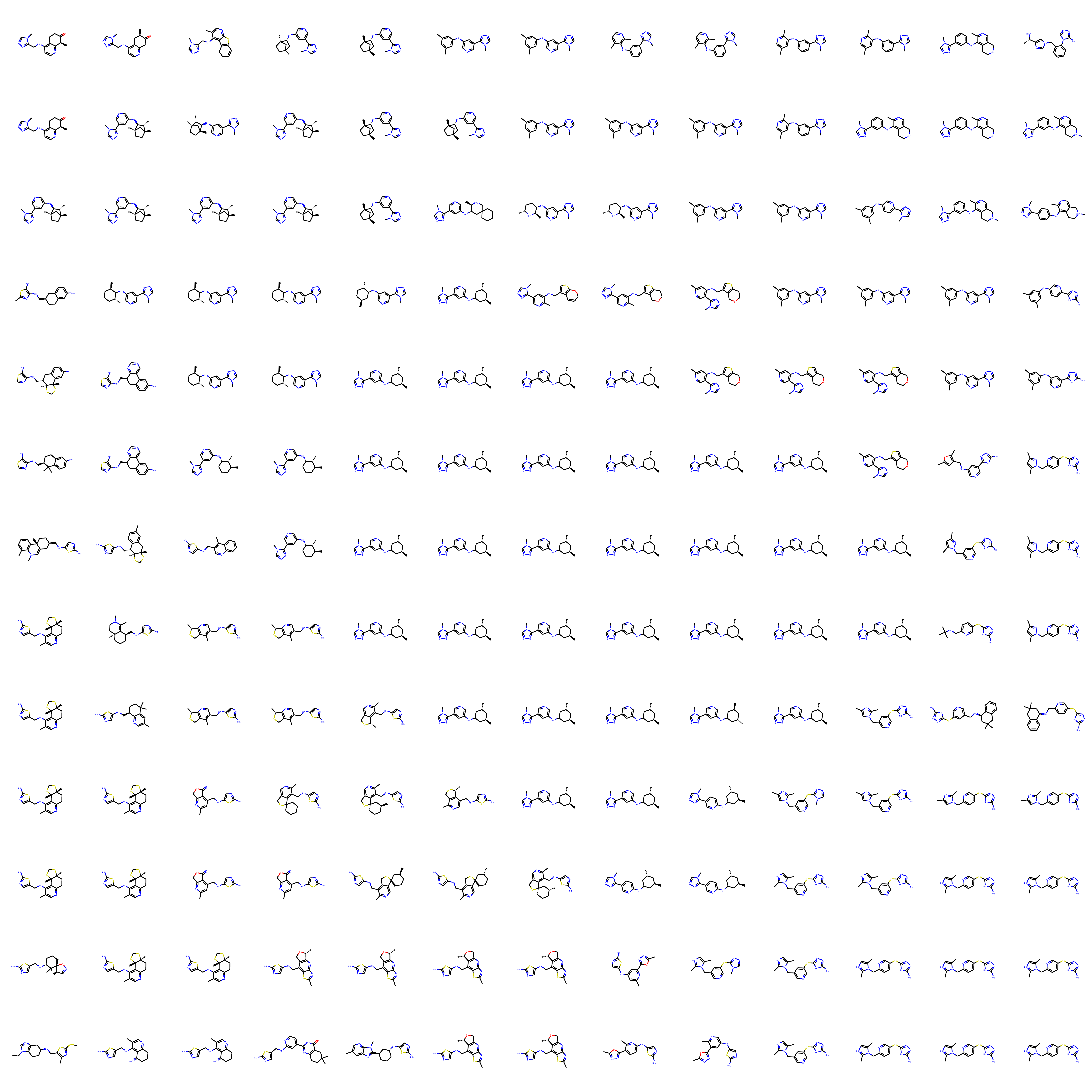}
\caption{Neighborhood visualization of molecule C[C@H]1CC(Nc2cncc(-c3nncn3C)c2)C[C@H](C)C1.}
\label{fig:neivis1}
\end{figure}

\begin{figure}[t]
\centering
\includegraphics[width=\textwidth]{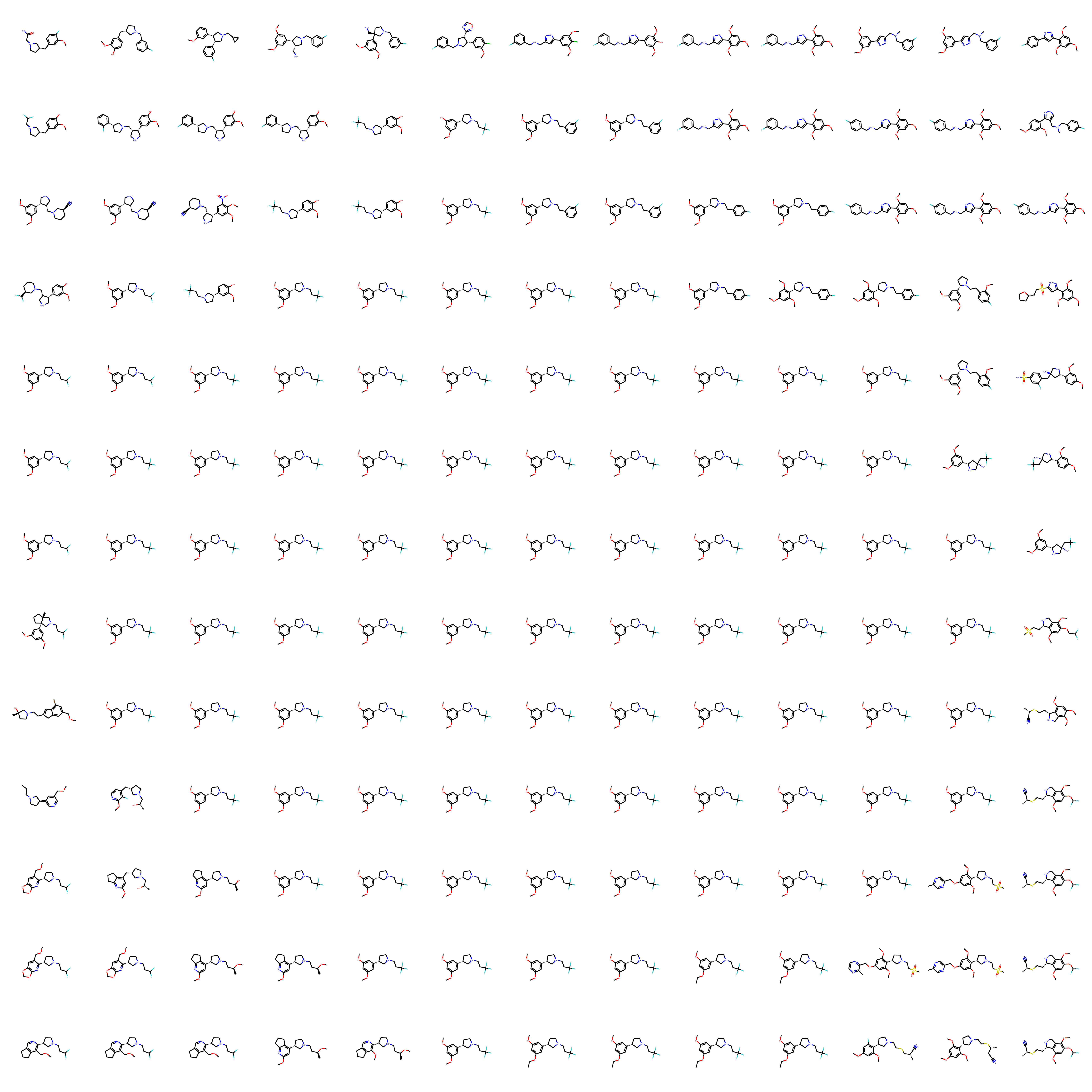}
\caption{Neighborhood visualization of molecule COc1cc(OC)cc([C@H]2CC[NH+](CCC(F)(F)F)C2)c1.}
\label{fig:neivis2}
\end{figure}

\textbf{Neighborhood Visualization } Given a molecule, we follow \citet{kusner2017grammar} to construct a grid visualization of its neighborhood. Specifically, we encode a molecule into the latent space and generate two random orthogonal unit vectors as two axis of a grid. Moving in combinations of these directions yields a set of latent vectors and we decode them into corresponding molecules. In Figure \ref{fig:neivis1} and \ref{fig:neivis2}, we visualize the local neighborhood of two molecules presented in \citet{dai2018syntax-directed}. Figure \ref{fig:neivis1} visualizes the same molecule in Figure~\ref{fig:vis}, but with wider neighborhood ranges.

\begin{figure}[t]
\centering
\includegraphics[width=\textwidth]{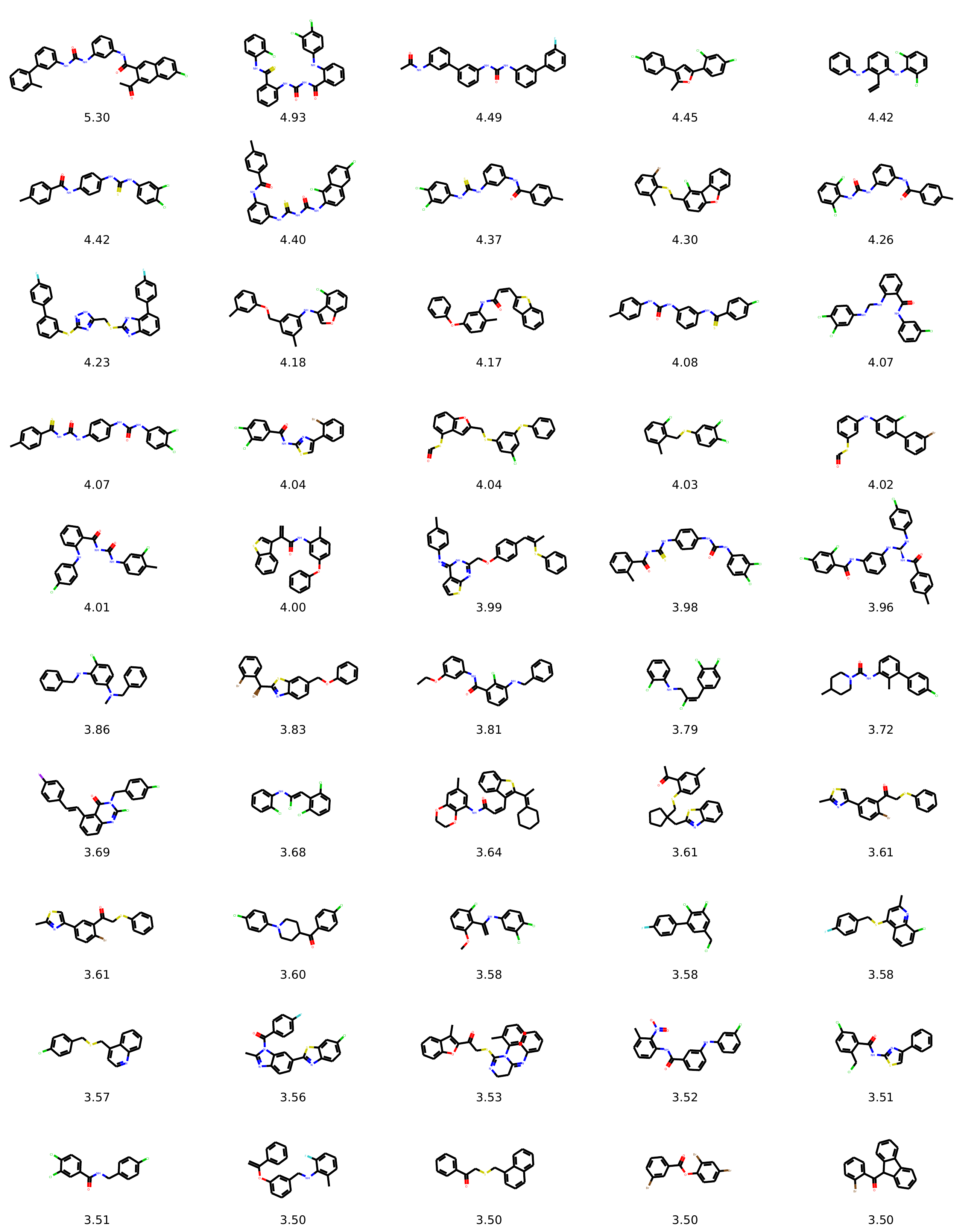}
\caption{Top 50 molecules found by Bayesian optimization using JT-VAE.}
\label{fig:bo-best50}
\end{figure}

\textbf{Bayesian Optimization } We directly used open sourced implementation in \citet{kusner2017grammar} for Bayesian optimization (BO). Specifically, we train a sparse Gaussian process with 500 inducing points to predict properties of molecules. Five iterations of batch BO with expected improvement heuristic is used to propose new latent vectors. In each iteration, 50 latent vectors are proposed, from which molecules are decoded and added to the training set for next iteration. We perform 10 independent runs and aggregate results. In Figure~\ref{fig:bo-best50}, we present the top 50 molecules found among 10 runs using JT-VAE. Following \citeauthor{kusner2017grammar}'s implementation, the scores reported are \textbf{normalized} to zero mean and unit variance by the mean and variance computed from training set.

\begin{figure}[t]
\centering
\includegraphics[width=\textwidth]{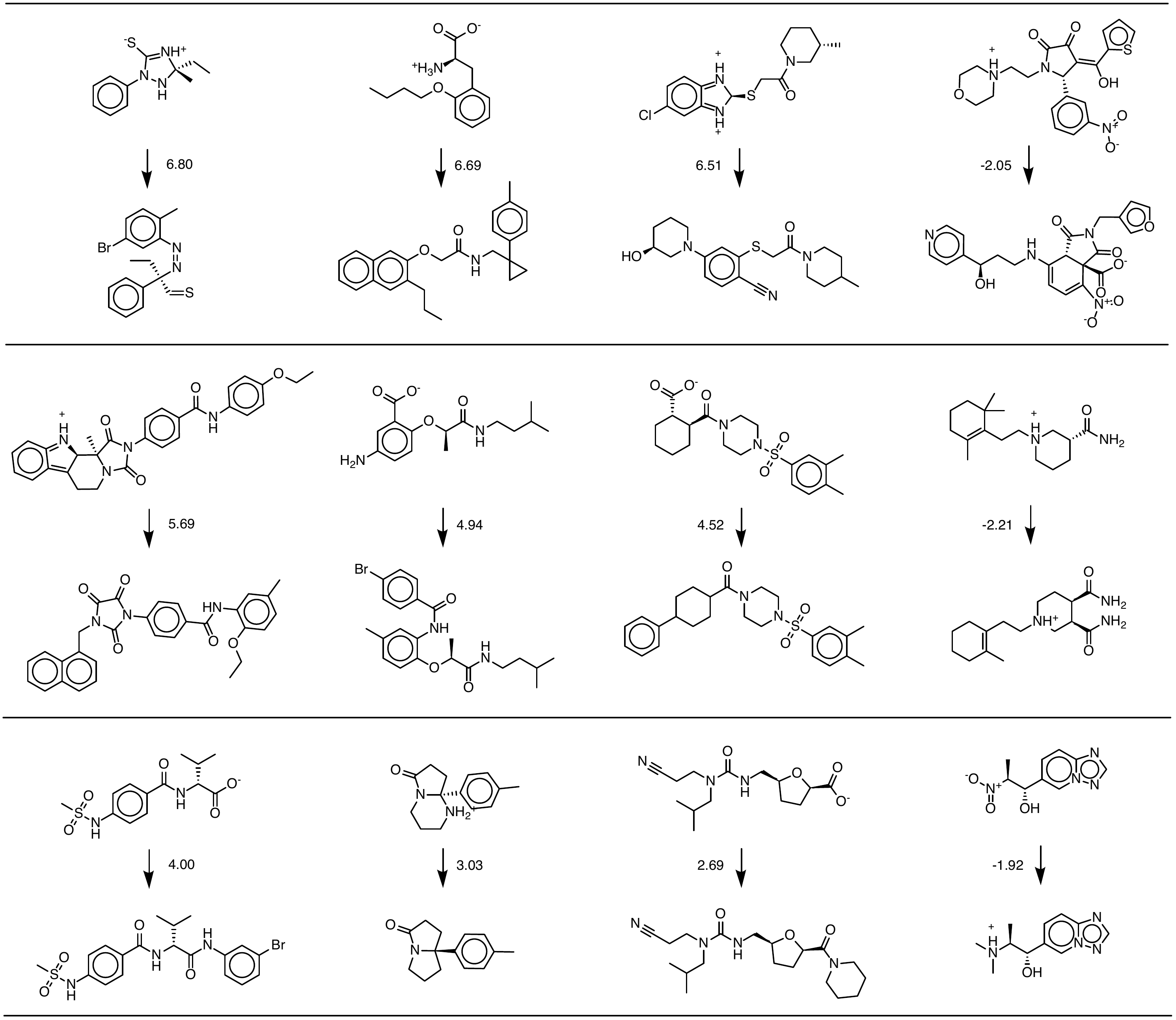}
\caption{Row 1-3: Molecule modification results with similarity constraint $sim(m,m') \geq 0.2,0.4,0.6$. For each group, we plot the top three pairs that leads to actual property improvement, and one pair with decreased property. We can see that tighter similarity constraint forces the model to preserve the original structure.}
\label{fig:all-molmodify}
\end{figure}

\textbf{Constrained Optimization } For this task, a property predictor $F$ is trained jointly with VAE to predict $y(m)=logP(m) - SA(m)$ from the latent embedding of $m$. $F$ is a feed-forward network with one hidden layer of dimension 450 followed by $\tanh$ activation. To optimize a molecule $m$, we start with its mean encoding $\z_m^0 = \bmu_m$ and apply 80 gradient ascent steps: $\z_m^t = \z_m^{t-1} + \alpha \frac{\partial y}{\partial z}$ with $\alpha=2.0$. 80 molecules are decoded from latent vectors $\{\z_m^i\}$ and their property is calculated. Molecular similarity $sim(m,m')$ is calculated via Morgan fingerprint of radius 2 with Tanimoto similarity. For each molecule $m$, we report the best modified molecule $m'$ with $sim(m,m')>\delta$ for some threshold $\delta$. In Figure~\ref{fig:all-molmodify}, we present three groups of modification examples with $\delta=0.2,0.4,0.6$. For each group, we present top three pairs that leads to best improvement $y(m')-y(m)$ as well as one pair decreased property ($y(m')<y(m)$). This is caused by inaccurate property prediction. From Figure~\ref{fig:all-molmodify}, we can see that tighter similarity constraint forces the model to preserve the original structure.

\end{document}